\documentclass[conference]{IEEEtran}
\IEEEoverridecommandlockouts
\usepackage{cite}
\usepackage{amsmath,amssymb,amsfonts}
\usepackage{algorithmic}
\usepackage{graphicx}
\usepackage{textcomp}
\usepackage{xcolor}

\usepackage{multirow}
\usepackage{booktabs}
\usepackage{graphics}
\usepackage{adjustbox}
\usepackage{subcaption}
\usepackage[font=small,labelfont=bf]{caption}

\def\BibTeX{{\rm B\kern-.05em{\sc i\kern-.025em b}\kern-.08em
    T\kern-.1667em\lower.7ex\hbox{E}\kern-.125emX}}
\begin{document}

\title{A New Approach to Improve Learning-based Deepfake Detection in Realistic Conditions \\
\thanks{The authors acknowledge support from CHIST-ERA project XAIface
(CHIST-ERA-19-XAI-011) with funding from the Swiss National Science
Foundation (SNSF) under grant number 20CH21 195532.}
}

\author{\IEEEauthorblockN{Yuhang Lu and  Touradj Ebrahimi}
\IEEEauthorblockA{\textit{Multimedia Signal Processing Group (MMSPG)} \\
\textit{\'Ecole Polytechnique F\'ed\'erale de Lausanne (EPFL)}\\
CH-1015 Lausanne, Switzerland \\
Email: firstname.lastname@epfl.ch}
}

\maketitle

\begin{abstract}
Deep convolutional neural networks have achieved exceptional results on multiple detection and recognition tasks. However, the performance of such detectors are often evaluated in public benchmarks under constrained and non-realistic situations. The impact of conventional distortions and processing operations found in imaging workflows such as compression, noise, and enhancement are not sufficiently studied. Currently, only a few researches have been done to improve the detector robustness to unseen perturbations. This paper proposes a more effective data augmentation scheme based on real-world image degradation process. This novel technique is deployed for deepfake detection tasks and has been evaluated by a more realistic assessment framework. Extensive experiments show that the proposed data augmentation scheme improves generalization ability to unpredictable data distortions and unseen datasets. 


\end{abstract}


\section{Introduction}
In recent years, deep convolutional neural networks (DCNN) trained with large-scale datasets have demonstrated significant improvements over most computer vision tasks, such as object detection, face recognition, forgery detection, etc. However, the performance of the DCNN-based methods are often assessed by specific public benchmarks under constrained situations that do not necessarily match reality. It has been shown that a dramatic performance deterioration could occur when facing unseen contents, perturbations or post-processing operations \cite{Dodge2016UnderstandingHI}, \cite{Zhou2017OnCO}, \cite{hendrycks2019robustness}. 
In realistic scenarios, image data can suffer from extrinsic distortions during acquisition, such as varying illumination, occlusion, noise, and afterwards even undergo pre- and post-processing operations in storage and delivery.
This could be a crucial problem for safety- and security-critical applications, such as surveillance, autonomous driving, and digital forensics, to mention a few examples. 

Face manipulation detection is used in many such applications. Deepfakes refer to manipulated face contents by deep learning tools. The development of such techniques and wide availability of open source software have simplified the creation of face manipulations, posing serious public concerns. While numerous deepfake detection methods have been published and have reported promising results on specific benchmarks, they have often been developed and assessed under unrealistic scenarios. A deployed detector could mistakenly block a pristine yet heavily compressed image. At the same time, a malicious agent could also fool the detector by simply adding imperceptible noise to fake media contents.
Therefore, an effective technique to improve the model robustness to realistic distortions and processing operations is desired. 



In this work the following contributions have been made.
\begin{itemize}
    \item Inspired by real-world data degradation process, a novel data augmentation scheme is proposed, which improves the generalization ability of deepfake detectors while at marginal loss of accuracy on original benchmark.
    \item A new assessment framework is introduced for generic learning-based detection systems, which rigorously measures the robustness of a detector under more realistic situations.
    \item Extensive experiments and ablation studies have been conducted to evaluate the effectiveness of the augmentation scheme with the help of the framework.

\end{itemize}

\section{Related Work}


\textbf{Robustness benchmark}
In recent years, research has been conducted to explore the robustness of CNN-based methods towards real-world image corruptions. Dodge and Karam \cite{Dodge2016UnderstandingHI} measured the performance of image classification models with data disturbed by noise, blurring, and contrast changes. In \cite{hendrycks2019robustness}, Hendrycks et al. presented a corrupted version of ImageNet \cite{5206848} to benchmark the robustness of image recognition models against common image manipulations. \cite{michaelis2019dragon}, \cite{2020}, \cite{Sakaridis_2021_ICCV} focused on a safety-critical task, autonomous driving, and provided robustness benchmark for various relevant vision tasks, such as object detection and semantic segmentation.
However, current efforts have put more emphasis on corruptions from extrinsic environment and often focus on only one specific task. In this work, we introduce an assessment framework that is suitable for multiple detection tasks and additionally considers the impact of realistic processing operations in the real world.


\begin{figure*}[ht]
	\centering
	\begin{adjustbox}{width=0.8\textwidth}
    \includegraphics[]{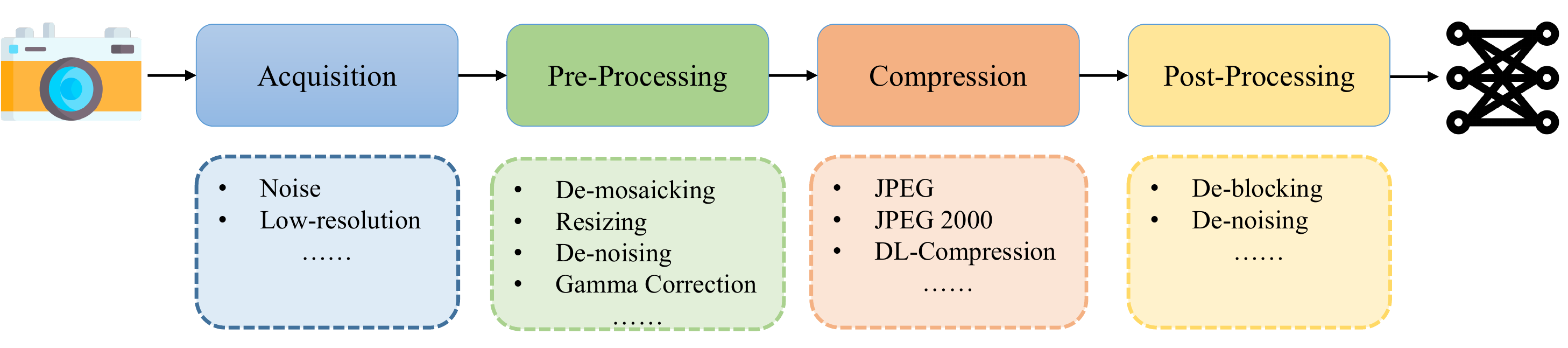}
	\end{adjustbox}
	\caption{A realistic data acquisition and transmission pipeline, including various natural image distortions and processing operations.}
	\label{fig:distortion}
\end{figure*}


\textbf{Face manipulation detection}
Face-centric detectors are often treated as a binary classification problem in computer vision. Early on, solutions based on facial expressions \cite{Agarwal_2019_CVPR_Workshops}, head movements \cite{Yang2019ExposingDF} and eye blinking \cite{9072088} were proposed. Current studies leverage deep learning technique to address such detection problems. Zhou et al. \cite{zhou_two-stream_2017} proposed to detect the deepfakes with a two-stream neural network. In \cite{roessler2019faceforensicspp}, R\"ossler et al. retrained an XceptionNet \cite{Chollet2017XceptionDL} with manipulated face dataset which outperforms on their proposed benchmark. Nguyen et al. \cite{Nguyen2019UseOA} combined traditional CNN and Capsule networks \cite{sabour2017dynamic}, which requires fewer parameters. Attention mechanisms have also been adopted to further improve the detection systems \cite{Zhao2021MultiattentionalDD}. Furthermore, to assist in a faster progress and better advancement of such tasks, numerous large-scale datasets, benchmarks, and competitions \cite{roessler2019faceforensicspp}, \cite{Celeb_DF_cvpr20}, \cite{jiang2020deeperforensics10},\cite{DFDC2020} have been made publicly available. For instance, FaceForensics++ \cite{roessler2019faceforensicspp} is one of the most popular face forensics datasets. Recently, Meta and Microsoft also organized a well-known Deepfake Detection Challenge based on their released dataset, DFDC \cite{DFDC2020}.  

\textbf{Data augmentation against corruption} Data augmentation is a popular technique in deep learning-based approaches, which often includes operations like flipping, cropping, translation, etc. Many studies leverage data augmentation technique to improve the generalization ability and to reduce the negative impact of natural distortions. Mixup \cite{zhang2018mixup} linearly combined two images and corresponding labels in the training batch. AugMix \cite{hendrycks2019augmix} utilized stochasticity and cascaded various augmentation operations and achieved the state-of-the-art on ImageNet-C \cite{hendrycks2019robustness}.
\cite{Ford2019AdversarialEA}, \cite{rusak2020increasing} have explored augmenting training data with Gaussian noise and managed to improve the performance of a object classifier on corrupted images.




\begin{table*}[t]
  \centering
  \caption{AUC (\%) scores of two detectors tested on unaltered and distorted variants of FFpp and Celeb-DFv2 test set respectively. Capsule-Forensics detector is shortened as \textit{Capsule}. The suffixes \textit{Raw} and \textit{Full} refer to different quality settings of FFpp. \textit{DL-Comp} is shortened from deep learning-based compression. DnCNN \cite{Zhang2017BeyondAG} refers to a learning-based denoising approach.}
    \begin{adjustbox}{width=\textwidth}

    \begin{tabular}{ccccccccccccccccccccccccc}
    \toprule
    \multirow{2}[4]{*}{Methods} & \multirow{2}[4]{*}{TrainSet} & \multirow{2}[4]{*}{Unaltered} & \multicolumn{3}{c}{JPEG} & \multicolumn{3}{c}{DL-Comp} & \multicolumn{3}{c}{Gau Noise} & \multirow{2}[4]{*}{\shortstack{Pois-Gau \\ Noise}} & \multicolumn{3}{c}{Gau Blur} & \multicolumn{3}{c}{Gamma Corr} & \multicolumn{3}{c}{Resize} & \multicolumn{3}{c}{Gau Noise+DnCNN} \\
\cmidrule{4-12}\cmidrule{14-25}          &       &       & 95    & 60    & 30    & High  & Med   & Low   & 5     & 30    & 50    &       & 3     & 7     & 11    & 0.1   & 0.75  & 2.5   & x4    & x8    & x16   & 10    & 30    & 50 \\
    \midrule
    \midrule
    \multirow{5}[4]{*}{Capsule} & FFpp-Raw & 99.20 & 97.91 & 76.48 & 59.60 & 44.76 & 45.50 & 49.08 & 61.80 & 51.26 & 49.16 & 55.63 & 67.19 & 41.78 & 47.74 & 49.50 & 98.86 & 96.12 & 67.48 & 47.82 & 46.90 & 75.09 & 66.68 & 60.63 \\
          & FFpp-Full & 94.52 & 94.95 & 93.97 & 84.50 & 86.83 & 60.98 & 55.69 & 89.03 & 57.95 & 51.11 & 64.87 & 85.72 & 58.83 & 56.05 & 56.02 & 93.86 & 85.44 & 87.05 & 69.93 & 54.15 & 86.24 & 80.54 & 73.85 \\
          & FFpp-Raw+Aug & 98.16 & 97.97 & 96.36 & 94.08 & 93.81 & 71.41 & 59.74 & 97.05 & 83.51 & 75.09 & 90.04 & 96.86 & 90.32 & 80.31 & 60.17 & 97.68 & 96.91 & 93.54 & 79.22 & 58.05 & 90.61 & 84.32 & 77.26 \\
\cmidrule{2-25}          & CelebDFv2 & 99.14 & 99.32 & 98.88 & 93.07 & -      & -       & -      & 95.24 & 59.32 & 60.57 & -      & 99.01 & 91.04 & 77.52 & 79.6  & 98.52 & 94.62 & 89.22 & 66.98 & 61.94 & 94.98 & 85.70 & 75.50 \\
          & CelebDFv2+Aug & 98.20 & 98.20 & 97.68 & 96.01 & -      & -      & -     & 96.88 & 83.51 & 74.71 & -      & 97.53 & 94.57 & 88.72 & 81.84 & 98.02 & 96.85 & 94.04 & 79.91 & 61.51 & 96.00 & 92.12 & 86.19 \\
    \midrule
    \multirow{4}[4]{*}{XceptionNet} & FFpp-Raw & 99.56 & 76.77 & 56.00 & 54.20 & 50.16 & 50.37 & 50.10 & 50.12 & 49.64 & 49.30 & 48.98 & 68.76 & 55.61 & 50.70 & 54.66 & 98.66 & 70.45 & 68.60 & 55.80 & 50.45 & 63.77 & 57.95 & 55.92 \\
          & FFpp-Raw+Aug & 98.44 & 98.25 & 97.36 & 96.12 & 98.03 & 87.76 & 82.74 & 97.37 & 91.71 & 88.70 & 94.57 & 98.31 & 97.35 & 94.51 & 80.48 & 98.25 & 97.75 & 97.30 & 86.26 & 67.14 & 94.18 & 90.68 & 86.95 \\
\cmidrule{2-25}          & CelebDFv2 & 99.73 & 99.59 & 99.78 & 97.75 & -      & -       & -      & 94.85 & 52.49 & 52.50 & -      & 98.77 & 91.81 & 79.94 & 74.98 & 99.53 & 97.49 & 96.01 & 72.20 & 63.03 & 97.81 & 91.44 & 80.29 \\
          & CelebDFv2+Aug & 99.01 & 99.00 & 98.80 & 98.24 & -      & -      & -      & 98.80 & 95.86 & 92.46 & -      & 98.84 & 97.18 & 95.03 & 72.53 & 98.89 & 98.11 & 97.60 & 89.08 & 73.59 & 97.80 & 95.83 & 92.44 \\
    \bottomrule
    \end{tabular}%
    \end{adjustbox}
  \label{tab:result}%
\end{table*}%


\section{Improved Training Scheme by Data Augmentation}

To reduce the negative impact of realistic distortions and post-processing operations on detection performance, we propose a simple but effective data augmentation approach that leads to a robustness improvement.


Many studies have explored using corrupted data as augmentations by adding Gaussian noise or applying compression to training data \cite{Ford2019AdversarialEA}, \cite{rusak2020increasing}, or transferring the data style \cite{michaelis2019dragon}. 
Despite that such techniques have proven to enhance the robustness, they are usually inadequate for more complex real-world conditions and do not bring desired levels of improvements. 

This work is motivated by a typical data acquisition and transmission pipeline in real world, as shown in Figure \ref{fig:distortion}. Before being used for vision tasks, image data often suffers from natural distortions during acquisition followed by a set of processing operations. Based on the observation of data degradation process, a carefully designed augmentation chain was conceived, which produces augmented training data that are much closer to realistic conditions. 

Generally, the brightness and contrast of input image $x$ are first modified by image enhancement operator \textbf{\textit{enh}}. Afterwards, the image is convoluted with an image blurring kernel \textbf{\textit{f}}, followed by additive Gaussian noise \textbf{\textit{n}}. At the end, \textbf{\textit{JPEG}} compression is applied to obtain the augmented training data $x_{\text{aug}}$. 
The augmentation chain is described by the following formula.

\begin{equation}
x_{\text{aug}} = \textbf{\textit{JPEG}}[(\textbf{\textit{enh}}(x)\circledast \textbf{\textit{f}})+\textbf{\textit{n}}]
\end{equation}

Moreover, directly cascading several augmentation operations could result in a reduction of performance on the original benchmark. The proposed augmentation technique is implemented in a stochastic manner, i.e. each operation occurs with a certain probability and randomness level, which also better reflects the natural data degradation process. 
Figure \ref{fig:aug-example} shows a few samples of typical augmented data. 

%




\newcommand\w{0.25\linewidth}
\newcommand\y{\linewidth}
\begin{figure}[t]
\centering
\begin{subfigure}[b]{\w}
  \includegraphics[width=\y]{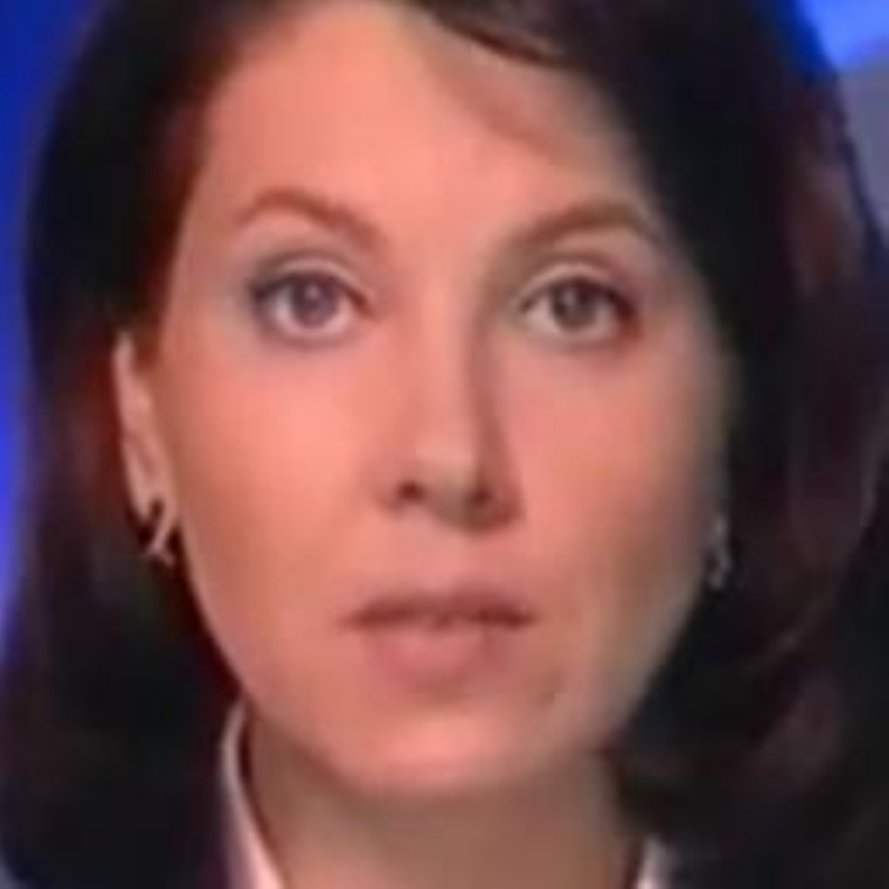}  
  \caption{Unaltered}
\end{subfigure}%
\hfill
\begin{subfigure}[b]{\w}
  \includegraphics[width=\y]{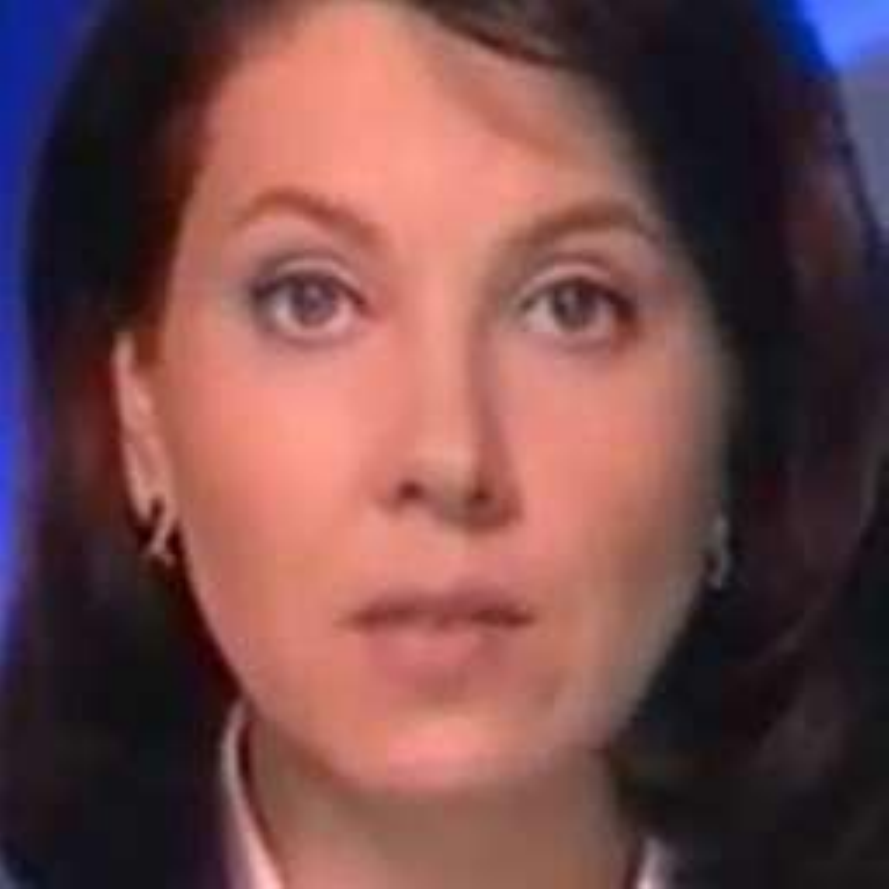}  
  \caption{JPEG}
\end{subfigure}%
\hfill
\begin{subfigure}[b]{\w}
  \includegraphics[width=\y]{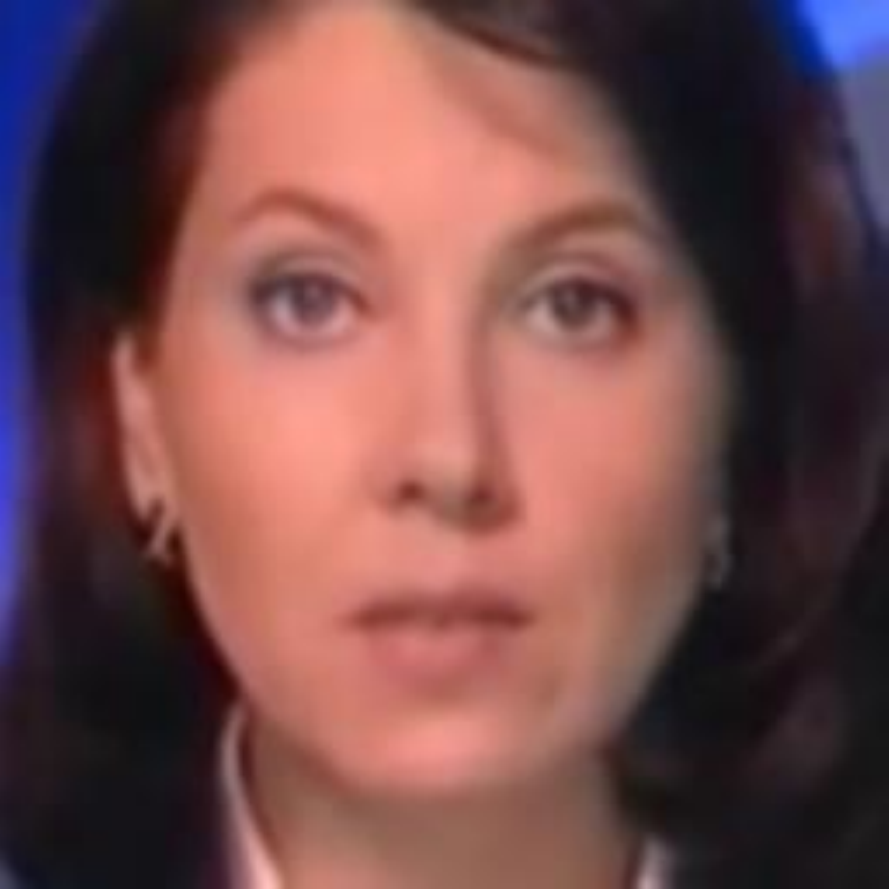}  
  \caption{GB}
\end{subfigure}%
\hfill
\begin{subfigure}[b]{\w}
  \includegraphics[width=\y]{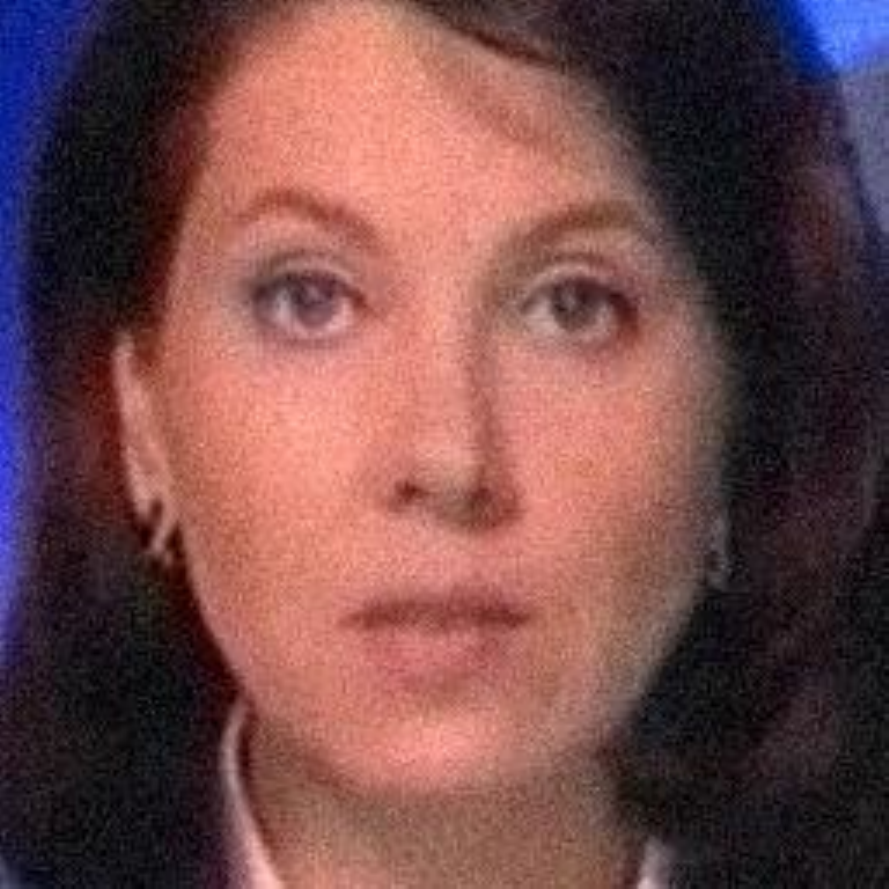}  
  \caption{GN}
\end{subfigure}%
\hfill
\begin{subfigure}[b]{\w}
  \includegraphics[width=\y]{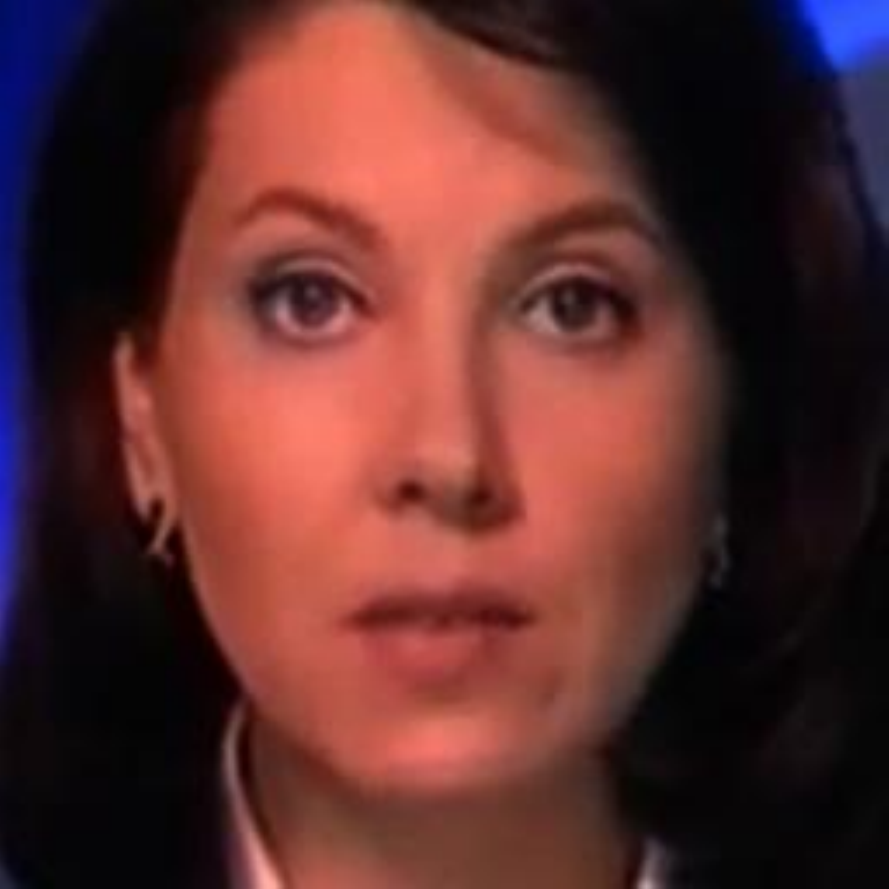}  
  \caption{GC}
\end{subfigure}%
\hfill
\begin{subfigure}[b]{\w}
  \includegraphics[width=\y]{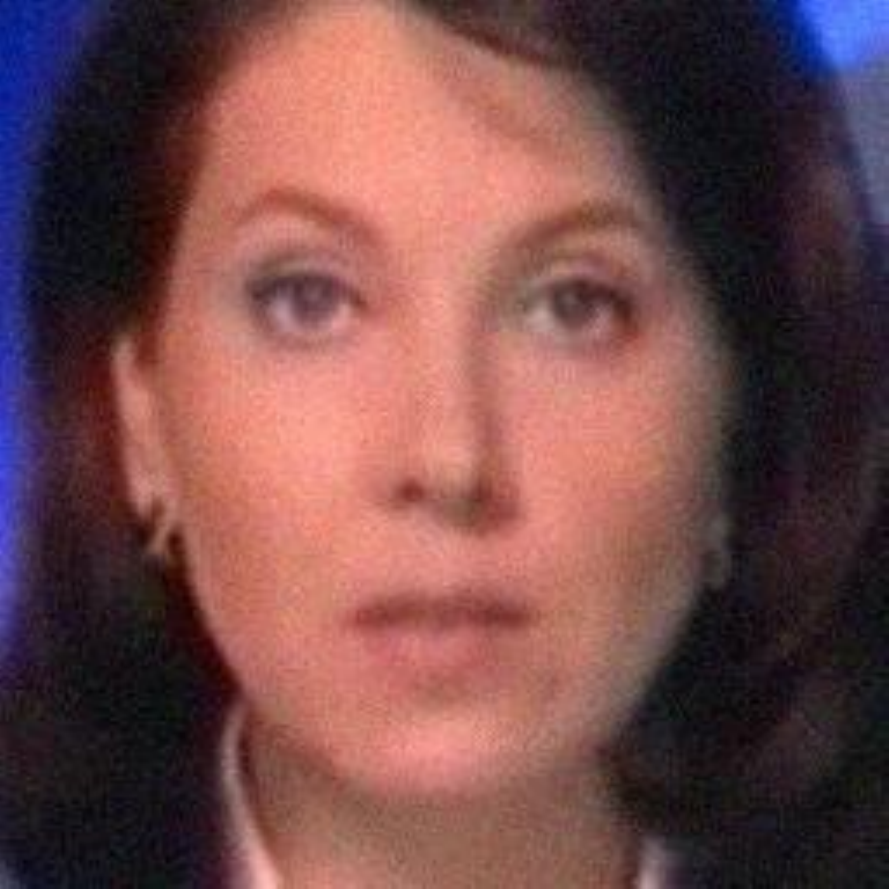}  
  \caption{GN+GB}
\end{subfigure}%
\hfill
\begin{subfigure}[b]{\w}
  \includegraphics[width=\y]{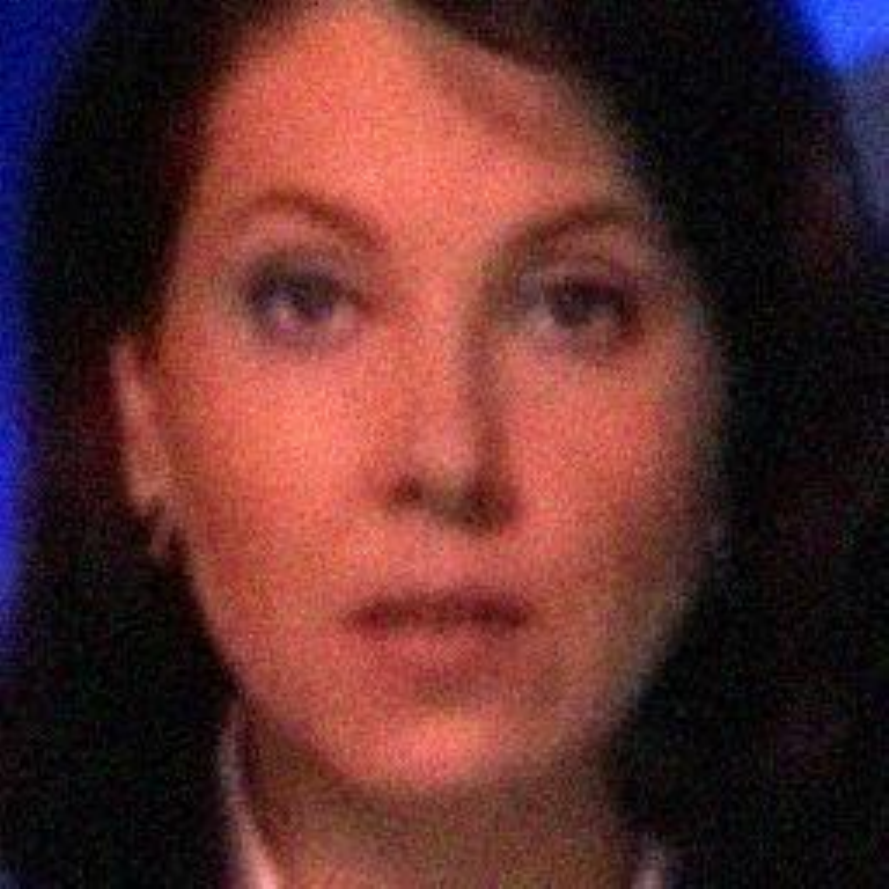}
  \caption{GB+GN+GC}
\end{subfigure}%
\hfill
\begin{subfigure}[b]{\w}
  \includegraphics[width=\y]{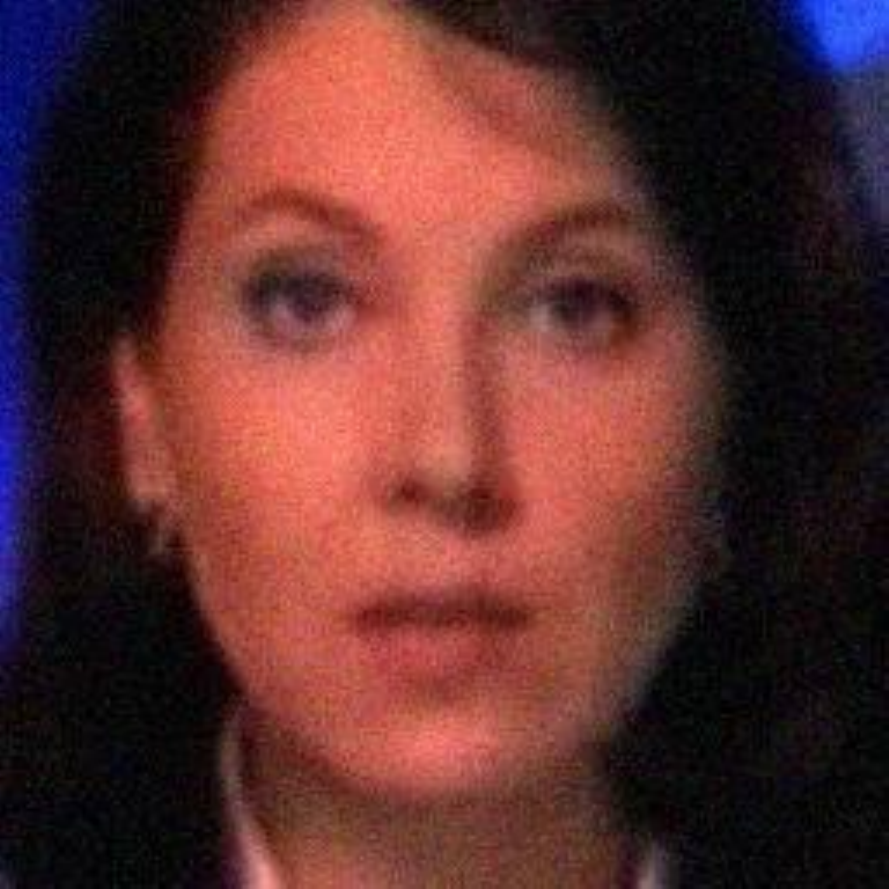}  
  \caption{All}
\end{subfigure}%

\caption{Example of some typical frames in the training dataset after applying the stochastic data augmentation scheme. Some notations are explained as following. GB: Gaussian blur. GN: Gaussian noise. GC: Gamma correction. $+$: mixture. All: all the operations cascaded.}
\label{fig:aug-example}
\end{figure}


In detail, the augmentation operations are explained in sequence as follows. 

\textit{Enhancement}: The augmentation chain begins with image enhancement operation. A probability of 50\% is adopted to apply either a brightness or a contrast operation on the training data which will be then non-linearly modified by a factor randomly selected from $[0.5, 1.5]$. 

\textit{Smoothing}: Image blurring operation is then applied with a selected probability of 50\%. Either Gaussian blur or Average blur filter is used with a kernel size varying in the range $[3, 15]$. 

\textit{Additive Gaussian Noise}: For each batch of training data, a probability of 30\%  is adopted to add a  Gaussian noise. The standard deviation of the Gaussian noise varies randomly in the interval $[0, 50]$.

\textit{JPEG Compression }: Finally, JPEG compression is applied with a selected probability of 70\%. The  quality factor corresponding to the compression is randomly chosen in the the range $[10, 95]$. 


\section{Illustrative Example For Deepfake Detection}
The usage of the proposed augmentation strategy is illustrated in the context of deepfake detection , which has become a hot topic in digital forensics. 

\subsubsection{Detection Methods}
Experiments have been conducted with two deep learning-based deepfake detectors, both of which have reported excellent performance on popular benchmarks.  

\textbf{Capsule-Forensics} \cite{Nguyen2019UseOA} achieves high detection accuracy and meanwhile the network maintains a rather small amount of parameters by combining conventional CNN and Capsule network \cite{sabour2017dynamic}. 


\textbf{XceptionNet} \cite{Chollet2017XceptionDL} is a popular CNN architecture in many computer vision tasks. Ro\"ssler et al. first adopted it in \cite{roessler2019faceforensicspp} to detect face manipulations and achieved excellent performance in the FaceForensics++ benchmark on both compressed and uncompressed contents. 

\subsubsection{Datasets}
Two widely used face manipulation datasets are selected for extensive experimentations to demonstrate the effectiveness of the proposed augmentation technique. 

\textbf{FaceForensics++} \cite{roessler2019faceforensicspp}, denoted by FFpp, contains 1000 pristine and 4000 manipulated video generated by four different deepfake creation algorithms. Additionally, raw video contents are compressed with two quality parameters using the H.264 codec, dented as C23 and C40.
In our experiments, data of all quality levels are involved for training while only uncompressed data are used for the final assessment.  

\textbf{Celeb-DFv2} \cite{Celeb_DF_cvpr20} is another high quality dataset, with 590 pristine celebrity video and 5639 fake video. The test data is selected as recommended by \cite{Celeb_DF_cvpr20} while the training and validation set was split in 80\% and 20\% accordingly.


\subsubsection{Implementation Details} Both detectors were trained with Adam optimizer with $\beta_1=0.9$, $\beta_2=0.999$. The Capsule-Forensics model is trained from scratch for 25 epochs with a learning rate of $5 \times 10^{-4}$, and the XceptionNet model is fine tuned on the manipulated face datasets for 10 epochs with learning rate of $1 \times 10^{-3}$. For the datasets, 100 frames are randomly sampled from each video for training and 32 frames are extracted for testing. Face regions are detected and cropped using dlib toolbox \cite{dlib09}. The proposed data augmentation technique is then adopted during the training process.









\section{Experiment Results by Realistic Assessment Framework}

We demonstrate the effectiveness of our data augmentation strategy by evaluating it through a realistic assessment framework. 
In this section, the proposed assessment framework for generic detection and recognition tasks is first described. 
It provides a way to broadly benchmark robustness for different models towards realistic data distortions and processing operations, and at the same time providing insights on further improvement.
Afterwards, the evaluation results are presented and analyzed in-depth.

\subsection{Assessment Framework Details}
In the assessment framework, six major categories of processing operations or added distortions are included, which are further divided into more than ten minor types. 

\textit{Noise}: Noise is a typical natural distortion during image acquisition. The framework adopts zero-mean Gaussian noise with 6 levels of variance. A synthetic Poissonian-Gaussian noise \cite{4623175} is also considered in the framework to better reflect the realistic situations.  

\textit{Resizing}: Images captured in outdoor environment often have limited resolution and can dramatically reduce the performance of learning-based detectors \cite{article}, \cite{Li2019OnLF}. Thus, the impact of low-resolution images is measured.    

\textit{Compression}: Compression is a widely used processing operation before transmission. Both JPEG compression with multiple ratios and a deep learning-based compression technique developed by Ball\'e et al. \cite{balle2018variational} are considered in the framework.

\textit{Smoothing}: Image blurring is another commonly employed operation to reduce noise. The impact of three different filters with various kernel sizes and a learning-based denoising technique \cite{Zhang2017BeyondAG} is evaluated by the framework. 

\textit{Enhancement}: Image enhancement technique is frequently used to adjust image brightness and contrast for better display. We change the contrast and brightness of images by separately applying linear adjustment and Gamma correction.

\textit{Combinations}: A combination of two or three distortions or operations are considered, such as combining noise and compression, making the test data to better reflect realistic situations. 

To employ assessment framework, one detector should be trained on its original target datasets, such as FFpp for deepfake detection tasks. Processing operations and corruptions are only applied on testing data. 

\begin{figure}[t]
	\centering
    \includegraphics[width=\linewidth]{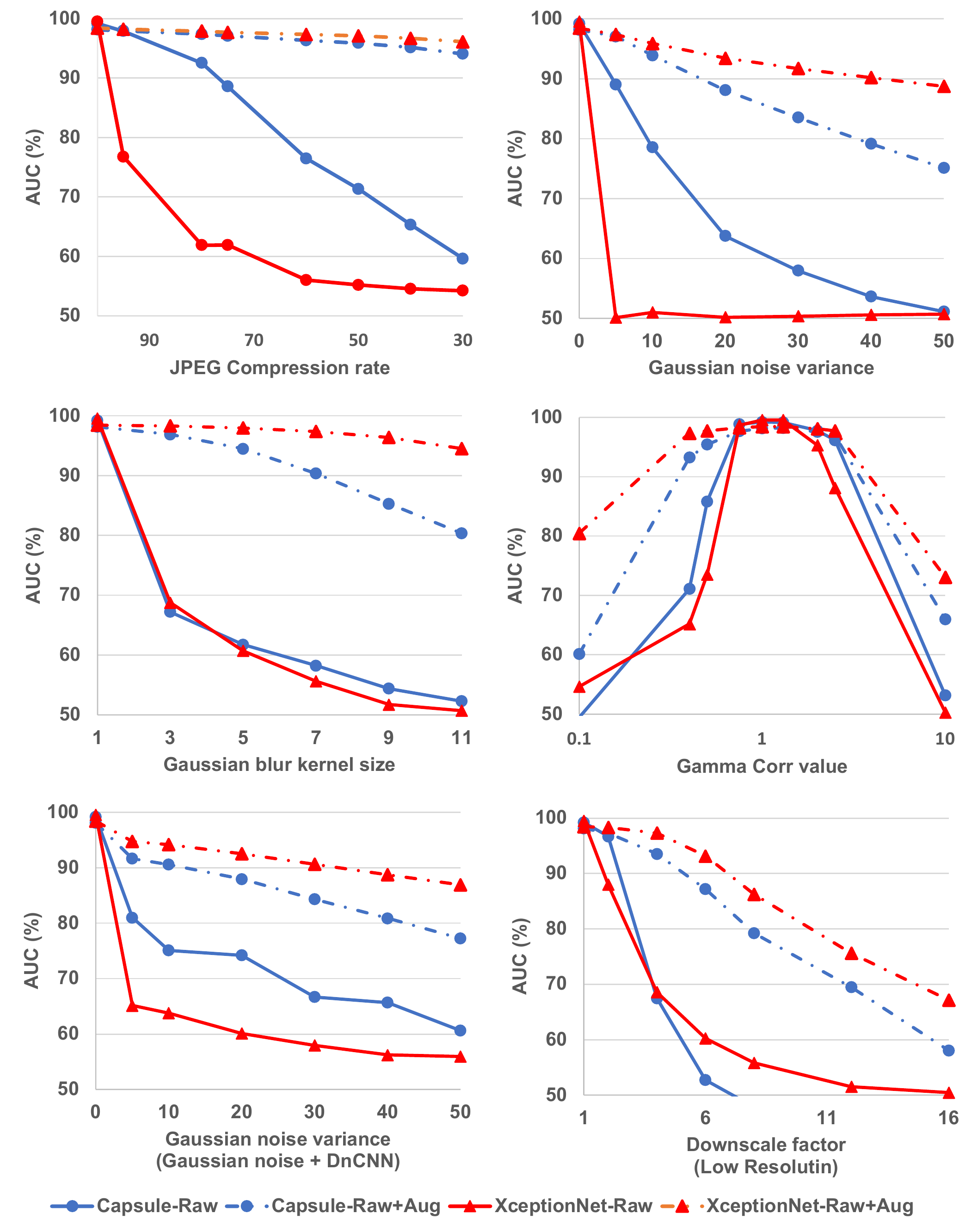}
	\caption{Performance comparison of models that are trained on FFpp-Raw only and trained with the proposed data augmentation scheme (+\textit{Aug}). Results of two different detectors are presented.}
	\label{fig:aug}
\end{figure}

\subsection{Experimental Results}
During the evaluation, we utilize Accuracy (ACC), the Area Under Receiver Operating Characteristic Curve (AUC), and F1-score in all experiments. 
In this section, only AUC scores and a subset of processing operations and severity levels are presented. 

The assessment results with the framework are presented in Table \ref{tab:result}. The information regarding the models trained with the proposed augmentation are denoted as +\textit{Aug}. In general, most of realistic distortions and processing are exceedingly harmful to normally trained learning-based deepfake detectors. For instance, Capsule-Forensics method shows very high AUC scores on both uncompressed FFpp and Celeb-DFv2 test set after training on respective datasets, but then suffers from drastic performance drop on modified data from our assessment framework. Similar trends have been observed with the XceptionNet detector. 

In comparison, it is evident that training with the proposed augmentation technique on the same dataset remarkably improves the performance on nearly all kinds of processed data even with intense severity. Figure \ref{fig:aug} further illustrates the impact of increasing the severity of four different types of realistic distortions and processing operations, i.e. JPEG compression, Gaussian noise, Gaussian blur, and Gamma correction. The data augmentation scheme significantly improves the robustness of the two detectors and meanwhile they still maintain high performance on original unaltered data. 

\begin{figure}[t]
	\centering
    \includegraphics[width=\linewidth]{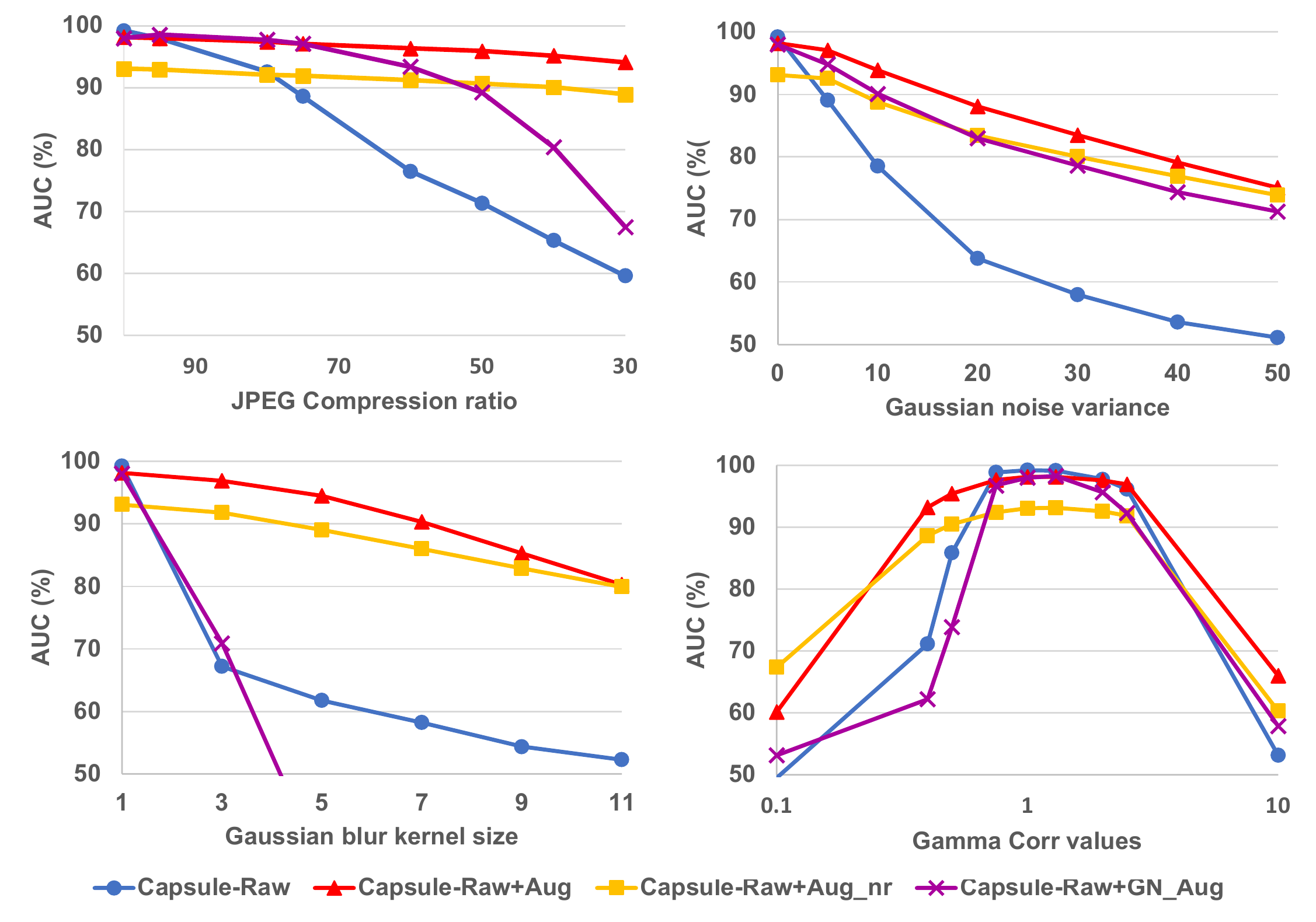}
	\caption{Frame-level AUC scores on four types of realistically distorted data after applying different augmentation strategy. Our proposed augmentation scheme is more robust to all corruptions.}
	\label{fig:ablation}
\end{figure}

Furthermore, as an ablation study, the effect of the cascaded augmentation chain as well as its randomized implementation is investigated. The results are presented in Figure \ref{fig:ablation}. We first train a Capsule-Forensics model on FFpp dataset augmented by only Gaussian noise, denoted as +\textit{GN\_Aug}. It is noticeable that the model only 
shows an improvement on noise corrupted data while still affected by other corruptions. 
The other model is trained with the same augmentation operations but without stochastic scheme, named +\textit{Aug\_nr}. As a result, there is an obvious performance decay on original unaltered benchmark. The reason for performance difference is that the proposed stochastic-based augmentation scheme can always maintain a certain amount of the original and less-distorted data for training.

\begin{table}[htbp]
  \centering
  \caption{Cross-dataset evaluation on Celeb-DFv2 (AUC(\%)) after training on FFpp dataset. The suffixes denote training set quality and the augmentation technique.}
    \begin{tabular}{ccccc}
    \toprule
    Method & TrainSet &       & FFpp  & Celeb-DFv2 \\
    \midrule
    \multirow{4}[2]{*}{Capsule} & FFpp-Raw &       & \textbf{99.20} & 54.39 \\
          & FFpp-C23 &       & 96.32 & 59.31 \\
          & FFpp-Full &       & 94.52 & 68.19 \\
          & FFpp-Raw+Aug &       & 97.82 & \textbf{71.86} \\
\cmidrule{1-2}\cmidrule{4-5}    \multirow{3}[2]{*}{XceptionNet} & FFpp-Raw &       & \textbf{99.56} & 50.00 \\
          & FFpp-Full &       & 99.02 & 64.58 \\
          & FFpp-Raw+Aug &       & 98.44 & \textbf{73.88} \\
    \bottomrule
    \end{tabular}%
  \label{tab:cross}%
\end{table}%



Finally, a cross-dataset assessment has been conducted to evaluate the generalization ability of the models trained with the proposed augmentation scheme on unseen datasets. The results are shown in Table \ref{tab:cross}. We train the selected detectors on FFpp dataset but test them on Celeb-DFv2 test set for frame-level AUC scores. The two methods both obtain relatively low scores on the new dataset, although mixing compressed data for training slightly improves the tranferability. The proposed augmentation scheme brings a certain performance improvement for both detectors on Celeb-DFv2, showing the capability to detect unseen forensic media contents.

\section{Conclusion}
Current detection methods are designed to achieve as high performance as possible on specific benchmarks. This often results in sacrificing generalization ability to more realistic scenarios. In this paper, a carefully conceived data augmentation scheme based on natural image degradation process is proposed. Extensive experiments show that the simple but effective technique significantly improves the model robustness against various realistic distortions and processing operations in typical imaging workflows.

\bibliographystyle{IEEEtran}
\bibliography{refs}

\begin{thebibliography}{10}
\providecommand{\url}[1]{#1}
\csname url@samestyle\endcsname
\providecommand{\newblock}{\relax}
\providecommand{\bibinfo}[2]{#2}
\providecommand{\BIBentrySTDinterwordspacing}{\spaceskip=0pt\relax}
\providecommand{\BIBentryALTinterwordstretchfactor}{4}
\providecommand{\BIBentryALTinterwordspacing}{\spaceskip=\fontdimen2\font plus
\BIBentryALTinterwordstretchfactor\fontdimen3\font minus
  \fontdimen4\font\relax}
\providecommand{\BIBforeignlanguage}[2]{{%
\expandafter\ifx\csname l@#1\endcsname\relax
\typeout{** WARNING: IEEEtran.bst: No hyphenation pattern has been}%
\typeout{** loaded for the language `#1'. Using the pattern for}%
\typeout{** the default language instead.}%
\else
\language=\csname l@#1\endcsname
\fi
#2}}
\providecommand{\BIBdecl}{\relax}
\BIBdecl

\bibitem{Dodge2016UnderstandingHI}
S.~F. Dodge and L.~Karam, ``Understanding how image quality affects deep neural
  networks,'' \emph{2016 Eighth International Conference on Quality of
  Multimedia Experience (QoMEX)}, pp. 1--6, 2016.

\bibitem{Zhou2017OnCO}
Y.~Zhou, S.~Song, and N.-M. Cheung, ``On classification of distorted images
  with deep convolutional neural networks,'' \emph{2017 IEEE International
  Conference on Acoustics, Speech and Signal Processing (ICASSP)}, pp.
  1213--1217, 2017.

\bibitem{hendrycks2019robustness}
D.~Hendrycks and T.~Dietterich, ``Benchmarking neural network robustness to
  common corruptions and perturbations,'' \emph{Proceedings of the
  International Conference on Learning Representations}, 2019.

\bibitem{5206848}
J.~Deng, W.~Dong, R.~Socher, L.-J. Li, K.~Li, and L.~Fei-Fei, ``Imagenet: A
  large-scale hierarchical image database,'' in \emph{2009 IEEE Conference on
  Computer Vision and Pattern Recognition}, 2009, pp. 248--255.

\bibitem{michaelis2019dragon}
C.~Michaelis, B.~Mitzkus, R.~Geirhos, E.~Rusak, O.~Bringmann, A.~S. Ecker,
  M.~Bethge, and W.~Brendel, ``Benchmarking robustness in object detection:
  Autonomous driving when winter is coming,'' \emph{arXiv preprint
  arXiv:1907.07484}, 2019.

\bibitem{2020}
\BIBentryALTinterwordspacing
C.~Kamann and C.~Rother, ``Benchmarking the robustness of semantic segmentation
  models,'' \emph{2020 IEEE/CVF Conference on Computer Vision and Pattern
  Recognition (CVPR)}, Jun 2020. [Online]. Available:
  \url{http://dx.doi.org/10.1109/CVPR42600.2020.00885}
\BIBentrySTDinterwordspacing

\bibitem{Sakaridis_2021_ICCV}
C.~Sakaridis, D.~Dai, and L.~Van~Gool, ``Acdc: The adverse conditions dataset
  with correspondences for semantic driving scene understanding,'' in
  \emph{Proceedings of the IEEE/CVF International Conference on Computer Vision
  (ICCV)}, October 2021, pp. 10\,765--10\,775.

\bibitem{Agarwal_2019_CVPR_Workshops}
S.~Agarwal, H.~Farid, Y.~Gu, M.~He, K.~Nagano, and H.~Li, ``Protecting world
  leaders against deep fakes,'' in \emph{Proceedings of the IEEE/CVF Conference
  on Computer Vision and Pattern Recognition (CVPR) Workshops}, June 2019.

\bibitem{Yang2019ExposingDF}
X.~Yang, Y.~Li, and S.~Lyu, ``Exposing deep fakes using inconsistent head
  poses,'' \emph{ICASSP 2019 - 2019 IEEE International Conference on Acoustics,
  Speech and Signal Processing (ICASSP)}, pp. 8261--8265, 2019.

\bibitem{9072088}
T.~Jung, S.~Kim, and K.~Kim, ``Deepvision: Deepfakes detection using human eye
  blinking pattern,'' \emph{IEEE Access}, vol.~8, pp. 83\,144--83\,154, 2020.

\bibitem{zhou_two-stream_2017}
P.~Zhou, X.~Han, V.~I. Morariu, and L.~S. Davis, ``Two-{Stream} {Neural}
  {Networks} for {Tampered} {Face} {Detection},'' in \emph{2017 {IEEE}
  {Conference} on {Computer} {Vision} and {Pattern} {Recognition} {Workshops}
  ({CVPRW})}, Jul. 2017, pp. 1831--1839, iSSN: 2160-7516.

\bibitem{roessler2019faceforensicspp}
A.~R\"ossler, D.~Cozzolino, L.~Verdoliva, C.~Riess, J.~Thies, and
  M.~Nie{\ss}ner, ``Face{F}orensics++: Learning to detect manipulated facial
  images,'' in \emph{International Conference on Computer Vision (ICCV)}, 2019.

\bibitem{Chollet2017XceptionDL}
F.~Chollet, ``Xception: Deep learning with depthwise separable convolutions,''
  \emph{2017 IEEE Conference on Computer Vision and Pattern Recognition
  (CVPR)}, pp. 1800--1807, 2017.

\bibitem{Nguyen2019UseOA}
H.~H. Nguyen, J.~Yamagishi, and I.~Echizen, ``Use of a capsule network to
  detect fake images and videos,'' \emph{ArXiv}, vol. abs/1910.12467, 2019.

\bibitem{sabour2017dynamic}
S.~Sabour, N.~Frosst, and G.~E. Hinton, ``Dynamic routing between capsules,''
  \emph{Advances in neural information processing systems}, vol.~30, 2017.

\bibitem{Zhao2021MultiattentionalDD}
H.~Zhao, W.~Zhou, D.~Chen, T.~Wei, W.~Zhang, and N.~Yu, ``Multi-attentional
  deepfake detection,'' \emph{2021 IEEE/CVF Conference on Computer Vision and
  Pattern Recognition (CVPR)}, pp. 2185--2194, 2021.

\bibitem{Celeb_DF_cvpr20}
Y.~Li, X.~Yang, P.~Sun, H.~Qi, and S.~Lyu, ``Celeb-df: A large-scale
  challenging dataset for deepfake forensics,'' in \emph{IEEE Conference on
  Computer Vision and Patten Recognition (CVPR)}, 2020.

\bibitem{jiang2020deeperforensics10}
L.~Jiang, R.~Li, W.~Wu, C.~Qian, and C.~C. Loy, ``Deeperforensics-1.0: A
  large-scale dataset for real-world face forgery detection,'' 2020.

\bibitem{DFDC2020}
B.~Dolhansky, J.~Bitton, B.~Pflaum, J.~Lu, R.~Howes, M.~Wang, and C.~C. Ferrer,
  ``The deepfake detection challenge dataset,'' 2020.

\bibitem{zhang2018mixup}
H.~Zhang, M.~Cisse, Y.~N. Dauphin, and D.~Lopez-Paz, ``mixup: Beyond empirical
  risk minimization,'' in \emph{International Conference on Learning
  Representations}, 2018.

\bibitem{hendrycks2019augmix}
D.~Hendrycks, N.~Mu, E.~D. Cubuk, B.~Zoph, J.~Gilmer, and B.~Lakshminarayanan,
  ``Augmix: A simple data processing method to improve robustness and
  uncertainty,'' in \emph{International Conference on Learning
  Representations}, 2019.

\bibitem{Ford2019AdversarialEA}
N.~Ford, J.~Gilmer, N.~Carlini, and E.~D. Cubuk, ``Adversarial examples are a
  natural consequence of test error in noise,'' in \emph{ICML}, 2019.

\bibitem{rusak2020increasing}
E.~Rusak, L.~Schott, R.~S. Zimmermann, J.~Bitterwolf, O.~Bringmann, M.~Bethge,
  and W.~Brendel, ``Increasing the robustness of dnns against image corruptions
  by playing the game of noise,'' 2020.

\bibitem{Zhang2017BeyondAG}
K.~Zhang, W.~Zuo, Y.~Chen, D.~Meng, and L.~Zhang, ``Beyond a gaussian denoiser:
  Residual learning of deep cnn for image denoising,'' \emph{IEEE Transactions
  on Image Processing}, vol.~26, pp. 3142--3155, 2017.

\bibitem{dlib09}
D.~E. King, ``Dlib-ml: A machine learning toolkit,'' \emph{Journal of Machine
  Learning Research}, vol.~10, pp. 1755--1758, 2009.

\bibitem{4623175}
A.~Foi, M.~Trimeche, V.~Katkovnik, and K.~Egiazarian, ``Practical
  poissonian-gaussian noise modeling and fitting for single-image raw-data,''
  \emph{IEEE Transactions on Image Processing}, vol.~17, no.~10, pp.
  1737--1754, 2008.

\bibitem{article}
T.~Marciniak, A.~Chmielewska, R.~Weychan, M.~Parzych, and A.~Dabrowski,
  ``Influence of low resolution of images on reliability of face detection and
  recognition,'' \emph{Multimedia Tools and Applications}, vol.~74, 06 2013.

\bibitem{Li2019OnLF}
P.~Li, L.~Prieto, D.~Mery, and P.~J. Flynn, ``On low-resolution face
  recognition in the wild: Comparisons and new techniques,'' \emph{IEEE
  Transactions on Information Forensics and Security}, vol.~14, pp. 2000--2012,
  2019.

\bibitem{balle2018variational}
J.~Ball{\'e}, D.~Minnen, S.~Singh, S.~J. Hwang, and N.~Johnston, ``Variational
  image compression with a scale hyperprior,'' in \emph{International
  Conference on Learning Representations}, 2018.

\end{thebibliography}


\end{document}